%% file: egpaper_final.tex
\documentclass[10pt,twocolumn,letterpaper]{article}
\pdfoutput=1 
\usepackage{iccv}
\usepackage{times}
\usepackage{epsfig}
\usepackage{graphicx}
\usepackage{amsmath}
\usepackage{amssymb}

\usepackage{booktabs}
\usepackage{graphics}
\usepackage{multirow}

\usepackage[pagebackref=true,breaklinks=true,letterpaper=true,colorlinks,bookmarks=false]{hyperref}

\iccvfinalcopy 

\def\iccvPaperID{2640} 

\ificcvfinal\fi

\begin{document}

\title{Learning Unified Decompositional and Compositional NeRF for \\ Editable Novel View Synthesis}

\author{Yuxin Wang$^1$ \and Wayne Wu$^2$ \and Dan Xu$^1$\thanks{Corresponding author} 
\vspace{3pt}
\and
$^1$Hong Kong University of Science and Technology\quad 
$^2$Shanghai Artificial Intelligence Laboratory \\
{\tt\small ywangom@connect.ust.hk, wuwenyan0503@gmail.com, danxu@cse.ust.hk}
}

\maketitle
\ificcvfinal\thispagestyle{empty}\fi

\input{00_abstract}
\input{01_intro}

\input{02_related}
\input{03_method}

\input{04_experiments}

\input{10_conclusion}

{\small

\input{egpaper_final.bbl}
\bibliographystyle{ieee_fullname}
}

\clearpage
\input{12_appendix}

\end{document}

%% file: 00_abstract.tex
\begin{abstract}
 Implicit neural representations have shown powerful capacity in modeling real-world 3D scenes, offering superior performance in novel view synthesis. 
In this paper, we target a more challenging scenario, i.e., joint scene novel view synthesis and editing based on implicit neural scene representations. 
State-of-the-art methods in this direction typically consider building separate networks for these two tasks (i.e., view synthesis and editing). Thus, the modeling of interactions and correlations between these two tasks is very limited, which, however, is critical for learning high-quality scene representations.
To tackle this problem, in this paper, we propose a unified Neural Radiance Field (NeRF) framework to effectively perform joint scene decomposition and composition for modeling real-world scenes. The decomposition aims at learning disentangled 3D representations of different objects and the background, allowing for scene editing, while scene composition models an entire scene representation for novel view synthesis.
Specifically, with a two-stage NeRF framework, we learn a coarse stage for predicting a global radiance field as guidance for point sampling, and in the second fine-grained stage, we perform scene decomposition by a novel one-hot object radiance field regularization module and 
a pseudo supervision via inpainting to handle ambiguous background regions occluded by objects. The decomposed object-level radiance fields are further composed by using activations from the decomposition module. 
Extensive quantitative and qualitative results show the effectiveness of our method for scene decomposition and composition, outperforming state-of-the-art methods for both novel-view synthesis and editing tasks\footnote{Project: \textcolor{magenta}{https://w-ted.github.io/publications/udc-nerf}}. 
\end{abstract}

%% file: 01_intro.tex
\vspace{-5pt}
\section{Introduction}
\label{sec:intro}

\begin{figure}
    \centering
    \includegraphics[width=0.95\linewidth]{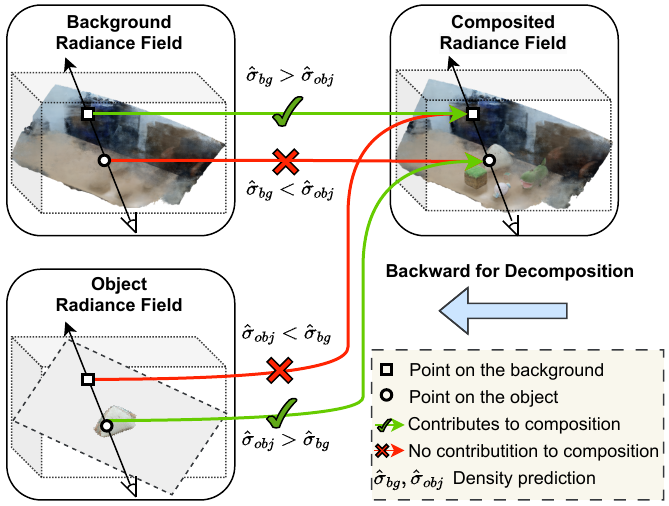}
    \vspace{-2pt}
    \caption{Illustration of our decomposition-composition design. It enables scene editing and novel view synthesis in a unified NeRF framework. The proposed decomposition targets learning disentangled 3D representations of different foreground objects and the background, allowing for scene editing, while scene composition constructs an entire scene representation for novel view synthesis. }
    \vspace{-18pt}
    \label{fig:teaser}
\end{figure}

The reconstruction and rendering of natural scenes are important for computers to understand the 3D physical world. Fine-grained object-level representations within the scene can bring significant benefits in various applications, such as scene understanding, novel content generation via object editing, and robotic manipulation. The emerging neural rendering techniques~\cite{mescheder2019occupancy,mildenhall2020nerf,park2019deepsdf} allow learning object-level or scene-level representations from multi-view posed images and enable rendering high-quality images from novel views. However, most implicit neural methods model an entire scene and lack fine-grained representations of objects in the scene, which severely limits object-level scene representation and understanding. 

\par 
Learning effective 3D object-aware implicit neural scene representations is still in its infancy. Existing works that tackle this challenging problem typically utilize additional object-level semantic supervision as a priori knowledge in the model optimization~\cite{wu2022objectsdf, yang2021objectnerf,Zhi:etal:ICCV2021}.
For instance, ObjectSDF~\cite{wu2022objectsdf} considers 
learning object-level geometries and semantics with an implicit signed distance function (SDF), which achieves effective scene composition from object-level representations. However, it remains challenging for this framework to perform joint object editing and novel view synthesis. On the other hand, ObjectNeRF~\cite{yang2021objectnerf} explores scene modeling with Neural Radiance Fields (NeRF) via introducing learnable object activation codes as switchers to condition the radiance field prediction among different objects. This framework can perform joint scene editing and novel view synthesis benefiting from the radiance field representations. However, it learns two separate networks for these two tasks, a global scene branch for novel view synthesis, and a local object branch for scene editing,  
without modeling any interaction or correlation among them, which, however, is critical for learning more effective scene representations, due to the simple fact that, global scene and local object representations are two important perspectives of the same scene. The global entire scene representation can model the scene's overall structures and appearance consistency, and the local object-specific representation can learn more fine-grained object details. They are highly complementary for learning a high-quality scene representation. 

\par To target the above-mentioned problem, this paper proposes a unified decompositional and compositional neural radiance field framework (see Fig.~\ref{fig:teaser}), to learn 3D scene representation for joint scene editing and novel view synthesis. 
The decomposition can provide the functionality of learning disentangled 3D representations of different objects and the background, allowing for scene editing, while scene composition models an entire scene representation for novel view synthesis. These two can be united to facilitate the consistency constraint in the unified optimization framework.
Specifically, we design a new two-stage framework for the scene decomposition and composition, consisting a coarse and a fine stage. In the coarse stage, we learn to predict a global scene radiance field as guidance for point sampling, and in the second fine-grained stage, we perform joint scene decomposition and composition. 
To perform effective scene decomposition, in the fine-stage, we apply a set of learnable object codes to predict distinct object-level radiance fields, and also propose two novel decomposition strategies: \textbf{(i)} 3D one-hot object radiance activation and regularization. For the object or background branches, only one branch is activated during training using a designed Gumbel-Softmax activation function directly applied on the point density predicted from different object/background branches with also a corresponding one-hot regularization; \textbf{(ii)} an in-painting pseudo supervision strategy. A challenge in scene rendering is modeling the appearance and geometry of regions occluded by objects. This causes generation ambiguity especially when the occluded regions are unseen in all the training views.  
To address this issue, we propose to use a pre-trained inpainting model to provide additional pseudo-color supervision for those ambiguous areas. 
Note that although the 2D inpainting may bring new ambiguity for the regions seen in other views, the supervision from multi-view consistency is strong enough to suppress most ambiguities. 
The composition is further performed by utilizing the learned one-hot activation weights for different object-level radiance fields.   

In summary, this paper has the following contributions:
\vspace{-3pt}
\begin{itemize}
\setlength{\itemsep}{0pt}
\setlength{\parskip}{2pt}
\item We propose a novel NeRF framework for joint scene decomposition and composition to effectively learn object-level and scene-level implicit representations for effective scene modeling, allowing object editing and novel view synthesis in a unified pipeline.
\item To learn a robust scene decomposition, we design two novel strategies, \textit{i.e.}, 3D one-hot radiance regularization and 2D in-painting pseudo supervision, to significantly improve the rendering 
 and editing qualities. 
\item 
Extensive experiments demonstrate the effectiveness of our approach for scene decomposition and show clear improvements upon state-of-the-art object-compositional methods on two important downstream tasks, \textit{i.e.}~novel-view synthesis and object editing.
\end{itemize}

%% file: 02_related.tex
\section{Related Work}
\label{sec:related}
\begin{figure*}[tp]
    \centering
    \includegraphics[width=1.0\linewidth]{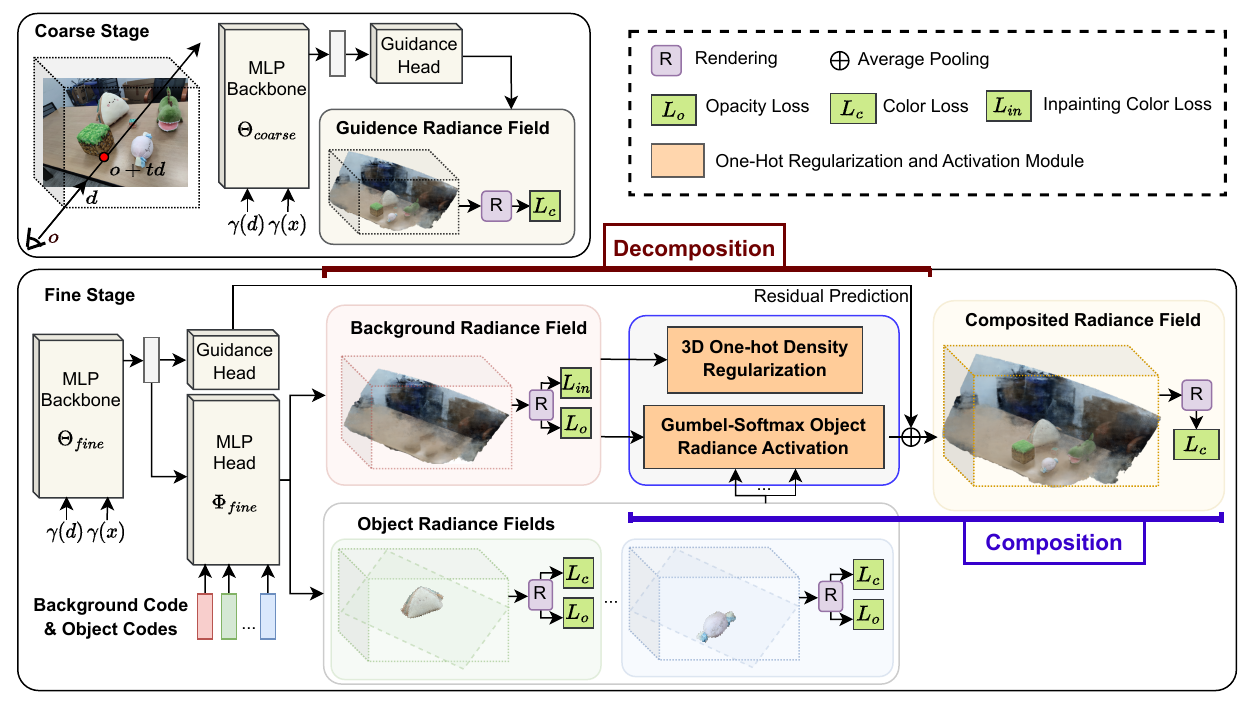}
    \caption{An overview of the proposed unified decompositional and compositional NeRF framework for joint novel view synthesis and scene editing. It has two stages. In the first stage (\ie~the coarse stage), it learns a guidance radiance field for guiding point sampling. In the second stage (\ie~the fine stage), we learn scene decomposition via learnable object codes and two novel decomposition schemes: (i) the 3D one-shot object radiance activation regularization and (ii) color inpaiting handling ambiguous generation in occluded background areas. The scene composition is achieved by using one-hot activation weights for different object-level radiance fields learned in the decomposition stage. The decomposition allows scene editing and the composition enables novel view synthesis in the unified framework.} 
    \vspace{-15pt}
    \label{fig:pipeline}
\end{figure*}

\par\noindent\textbf{Neural Implicit Representations.}
As pioneering methods,~\cite{mescheder2019occupancy,Niemeyer2020CVPR,park2019deepsdf,sitzmann2019srns} introduce neural networks to implicitly parameterize 3D scenes. These works can learn continuous shape and texture from differentiable rendering with~\cite{Local_Implicit_Grid_CVPR20,park2019deepsdf} or without\cite{Niemeyer2020CVPR,sitzmann2019srns} 3D supervision. 
Neural Radiance Fields (NeRF)~\cite{mildenhall2020nerf} 
optimize a 3D continuous neural scene representation by using only 2D multi-view posed images for training and produce high-quality novel view synthesis with 3D structure awareness. 
Recent works also extend the generalization of NeRF to more significant scenes~\cite{barron2021mipnerf,barron2022mip,kaizhang2020}, or model scenes with fewer perspectives~\cite{SRF,wang2021ibrnet,yu2021pixelnerf}. However, all these works only focus on the entire scene representation, neglecting fine-grained representations of objects/background within the scene. 
Other researches represent object surfaces with implicit neural function to realize multi-view 3D reconstruction for single objects~\cite{Oechsle2021ICCV,wang2021neus,yariv2021volume}.
The decomposed implicit representation for independently characterizing components within the scene has been studied in not only object scenes~\cite{benaim2022volumetricdisentanglement,kohli2020semantic,guo2020object,ost2021neural,rebain2021derf,srt22}, but also in the face or body tasks~\cite{guo2021adnerf,zhang2021stnerf,Wang_2021_CVPR}. In contrast to these works, our method targets both scene synthesis and editing, considering joint object- and scene-level implicit representation learning.

\par\noindent\textbf{Object-Compositional Implicit Representation.}
Representing scenes in an object-compositional manner can benefit both scene synthesis and editing~\cite{benaim2022volumetricdisentanglement,neural_outdoor_rerender,sajjadi2022osrt,smith2022unsupervised,stelzner2021decomposing,wu2022objectsdf,yu2022unsupervised,wu2023objsdfplus}. 
Compared to decomposed object representations represented by generative models~\cite{epstein2022blobgan,BlockGAN2020,Niemeyer2020GIRAFFE}, implicit models present more 3D physical meaning during the modeling process, which facilitates image rendering from novel views. 
We can categorize the works of object-compositional implicit representations into supervised and unsupervised methods. 
The unsupervised methods localize objects with different aspects of prior information, such as multiple similar scenes~\cite{sajjadi2022osrt,smith2022unsupervised,yu2022unsupervised}, temporal consistency~\cite{kipf2022conditional}, or clustering-based initialization~\cite{liu2022unsupervised}. Specifically, uORF~\cite{yu2022unsupervised} involves a slot-attention~\cite{kipf2022conditional} module in learning object-level iterative attention masks for each object decomposition. 
RFP~\cite{liu2022unsupervised} utilizes clustering predictions as initialization and applies a bidirectional color loss to decompose individual objects. 
However, lacking precise annotation, although these methods can obtain object position clues, they usually suffer from low-resolution~\cite{sajjadi2022osrt,yu2022unsupervised} issues that restrict generalization ability to high-resolution real-world scenes.
The supervised methods usually use object-level annotations, such as semantic masks~\cite{wu2022objectsdf, yang2021objectnerf, Zhi:etal:ICCV2021} or scene graphs~\cite{ost2021neural}. Among these works, 
ObjectNeRF~\cite{yang2021objectnerf} utilizes a two-branch network to learn a scene radiance field with an activation code-conditioned object radiance field supervised by multi-view images and semantic masks. 
Similarly, ObjectSDF~\cite{wu2022objectsdf} encodes the entire scene by combining the individual Signed Distance Functions (SDF) represented with semantic labels.
The most related work to us is ObjectNeRF~\cite{yang2021objectnerf}. In contrast to ObjectNeRF, we model an the scene representation via joint scene decomposition and composition, which can effectively model global consistency and rich local details in a unified NeRF framework for a high-quality scene representation.

%% file: 03_method.tex
\section{Decompositional and Compositional NeRF}
\label{sec:method}
\vspace{-3pt}
Given a set of multi-view posed images $\{ x_{i} , i\in [1, N] \}$ and their corresponding instance semantic masks $\{ s_{ij} , i\in [1, N], j\in [1, K+1] \}$ of a static real-world scene, our goal is to learn an entire scene representation based on the composition of a global scene radiance field and explicitly decomposed local radiance fields of different objects and the background. The decomposition allows scene editing and the composition enables novel view synthesis. $N$ is the number of multi-view images, and $K$ is the number of object instances, where we regard the background as an additional instance. We propose a joint decomposition and composition framework to allow simultaneous performing these two tasks. 

\subsection{Framework Overview}
An overall framework illustration is shown in Fig.~\ref{fig:pipeline}. 
It consists of two rendering stages. In the first stage (\ie~the coarse stage), it learns a global scene guidance radiance field for guiding point sampling. In
the second stage (\ie~the fine stage), we learn scene decomposition and composition. The decomposition is performed via first applying a set of learnable object codes as conditions to predict $K+1$ local object/background radiance field components, where $K$ is the number of objects, and the additional 1 represents the background. Following NeRF~\cite{mildenhall2020nerf}, we use continuous neural radiance fields to represent each local object/background. Considering the composition nature that the whole scene consists of all the $K+1$ components, we composite all the object and background radiance fields to an entire scene radiance field based on a proposed one-hot object radiance activation module. We also propose an effective color-inpainting strategy to handle the ambiguity in generating occluded background regions. The residual prediction from the first coarse stage is also utilized to benefit from a global scene representation. 
In the following subsections, we first give a general introduction to the proposed unified framework. Then, we introduce our novel decomposition and composition design and how the method can help disentangle different objects explicitly. 
After that, we present how our unified framework is optimized and utilized for both novel-view synthesis and editing tasks. 

\subsection{Scene Decomposition}
\noindent\textbf{Two-branch network architecture design.}  
For scene decomposition in the fine-stage, we design a two-branch network architecture with a shared MLP backbone to learn generic implicit scene representation from input point coordinates (see Fig.~\ref{fig:gumbel}).~Specifically, the shared backbone takes a sampled 3D point coordinate $\mathbf{p} = (x,y,z)$ and view direction $\mathbf{d}$ of a camera ray as input, and predicts a generic scene feature vector $\mathbf{f}$. Similar to~\cite{yang2021objectnerf}, we also apply a set of learnable codes, each representing the background or an object. We further pass $\mathbf{f}$ accompanying with background or object codes 
to the following MLP heads. 
For better background learning, we use separate the background head and the object head, predicting an RGB color $\mathbf{c}_{\text{object}}^{i}$ and a volume density $\sigma_{\text{object}}^{i}$ for the corresponding background or object $i$. 
For simplicity, we refer to the background as a particular object with index $i=1$. 
For efficient point sampling, we apply a coarse-to-fine sampling technique. In the coarse stage, we only adopt several MLPs after the backbone for predicting RGB color $\mathbf{c}_{\text{guidance}}$ and volume density $\sigma_{\text{guidance}}$ of the global scene. We use the distribution of coarse prediction for the fine point coordinate sampling, so we call these MLPs guidance head. In the fine stage, we also use the prediction of the guidance head as a residual radiance field in the following composition module.

\par\noindent\textbf{3D one-hot density regularization.}
To achieve effective decomposition of objects/background radiance fields from the scene, we also design a one-hot regularization term for the uncertain regions in the background radiance field. Our motivation is that the 3D points belonging to the foreground objects should not have any volume density in the background radiance field. We thus apply a one-hot regularization loss term, which acts as an additional constraint to the object-specific radiance field optimization objective $\mathcal{L}_{\text{object}}$. 
In detail, we first obtain the 3D point positions that foreground objects occupy by determining whether the foreground objects take the largest volume density at the points. We thus define a binary occupancy mask $M_c$, 
and apply a regularization term for the occupied points as follows:

\begin{equation}
\setlength{\abovedisplayskip}{-5pt}
\setlength{\belowdisplayskip}{6pt}
\begin{aligned}
& M_c =   [\text{argmax} ( [\sigma_{\text{object}}^{1}, \sigma_{\text{object}}^{2}, \sigma_{\text{object}}^{3}, ..., \sigma_{\text{object}}^{K+1}] ) > 1 ] \\
 & \mathcal{L}_{\text{one-hot}}= M_c \|\sigma_{\text{object}}^{1} - \sigma_{0} \|_2^2
\end{aligned}
\end{equation}
$\sigma_{\text{object}}^{1} $ here means we only want to constrain the background radiance field (with index 1). Note that we implement a ReLU operation on the volume density $\sigma$ to make it positive during the volume rendering process, so if the raw prediction $\sigma_{\text{object}}^{1}$ is negative at some points, it will also have no contribution to the rendering results. 
Therefore, we can set $\sigma_{0}$ as any non-positive number here. We put it -0.01 in our experiments from experience.

\par\noindent\textbf{2D in-painting pseudo supervision.}
To model a better background-specific radiance field, we need to learn what the scene looks like even if all the foreground objects are removed from the scene. Several occluded regions in the background can be seen in other views, while regions covered by the objects are invisible from all views, which are uncertain regions. We thus do not have any pixel-wise supervision for those regions in the training data, resulting in unsatisfactory extrapolation results on those regions when we learn the background radiance field.
To tackle this problem, we involve a pre-trained in-painting model to fill those uncertain regions on the 2D images. Our motivation is that the 2D in-painting model can utilize local context information on an image to fill the regions, providing effective 2D pseudo color supervision.
Specifically, we use a famous pre-trained lama ~\cite{suvorov2021resolution} model to fill the uncertain 2D background regions and then apply a color loss for the rendered background pixels with a {foreground mask} $M_1(\boldsymbol{r})^1$: 
\begin{equation}
\setlength\abovedisplayskip{8pt}
\begin{aligned}
\mathcal{L}_{\text{color\_pseudo}} = M_1(r)^1 \|\hat{C}(\boldsymbol{r})_{\text{object}}^{1}-C(\boldsymbol{r})_{\text{pseudo}}\|_2^2
\end{aligned}
\setlength\belowdisplayskip{6pt}
\end{equation}

\noindent\textbf{Intermediate local object rendering.} 
To enforce the decomposition of all object-specific radiance fields, we apply a color consistent loss and an opacity loss with semantic masks as ~\cite{mildenhall2020nerf,yang2021objectnerf} via rendering both 2D color and opacity:
\begin{equation}
\setlength\abovedisplayskip{4pt}
\begin{aligned}
\hat{C}(\boldsymbol{r})_{\text{object}}^{k} &=\sum_{i=1}^N T_i^k \big(1-\exp (-\sigma_{\text{object}}^{k} \delta_i) \big)\mathbf{c}_{\text{object}}^{k}, \\
\hat{O}(\boldsymbol{r})_{\text{object}}^{k} &=\sum_{i=1}^N T_i^k \big(1-\exp (-\sigma_{\text{object}}^{k}\delta_i)\big)
\end{aligned}
\setlength\belowdisplayskip{4pt}
\end{equation}
where $T_i^k =\exp (-\sum_{j=1}^{i-1} \sigma_{\text{object}}^{k} \delta_j)$ is the transmittance representing how much light is transmitted along the ray for the $k$-th object; $N$ is the number of sampled points along the camera ray, and $\delta_i$ is the distance between two adjacent points. Then we apply the following losses: 
\begin{equation}
\setlength\abovedisplayskip{5pt}
\begin{aligned}
\mathcal{L}_{\text {object}}=& \sum_{k=1}^{K+1} M_o(\boldsymbol{r})^k\|\hat{C}(\boldsymbol{r})_{\text{object}}^{k}-C(\boldsymbol{r})\|_2^2 \\
&+\lambda M_b(\boldsymbol{r})^1\|\hat{O}(\boldsymbol{r})_{\text{object}}^{1}-M_o(\boldsymbol{r})^1\|_2^2,
\end{aligned}
\setlength\belowdisplayskip{5pt}
\end{equation}
where $M_o(\boldsymbol{r})$ and $M_b(\boldsymbol{r})$ are 2D spatial weight maps determined by object-level and background-level semantic masks, respectively, which indicates that rendered object-specific pixels are given supervision at the corresponding semantic mask region. 
The background region in the binary semantic mask can be either real background or occluded by a foreground object, so we use a small weight value (\textit{i.e.} 0.05) in $M_b$ as a balance weight for the uncertain region. 

\subsection{Scene Composition}
This section presents how the learned global and the local object/background radiance fields are composited for an entire scene radiance field.
\par
\begin{figure}[tp]
    \centering
    \includegraphics[width=\linewidth]{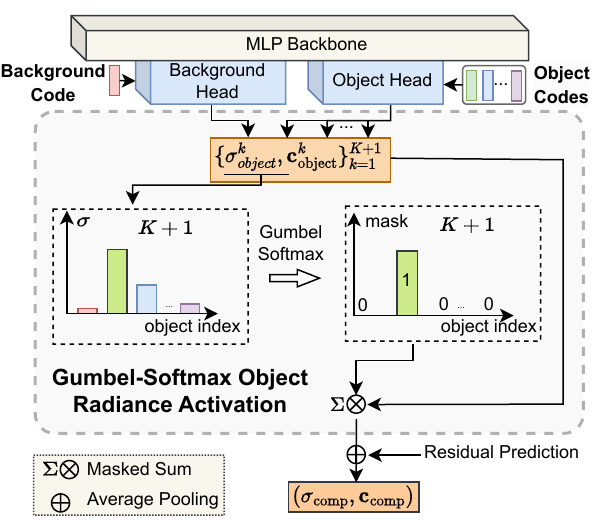}
    \caption{Illustration of the Composition Module. Our composition module consists of one-hot object radiance activation and global-local radiance field composition (the two green blocks). We first apply Gumbel-Softmax on object-level volume density values to obtain the one-hot mask for object radiance fields. Then, the local compositional radiance field is generated by a weighted sum of the object and background color and density predictions. Finally, the composited radiance is combined with a residual prediction to produce the final scene radiance field.}
    \vspace{-20pt}
    \label{fig:gumbel}
\end{figure}

\noindent\textbf{Gumbel-Softmax one-hot object radiance activation.} As shown in ~Fig.~\ref{fig:gumbel}, for each sampled point, there are $K+1$ predictions in the local radiance field decomposition. 
Considering that each point in the 3D space can only belong to either one object or the background, we use the predicted $(\mathbf{c}_{\text{object}}^{i}, \sigma_{\text{object}}^{i})$ with the maximum volume density $\sigma_{\text{object}}^{i}$ among all the object instances to construct an activated object-specific radiance field. More specifically, to address the gradient backpropagation problem of the argmax operation, following ~\cite{DBLP:conf/iclr/JangGP17,DBLP:conf/iclr/MaddisonMT17}, We employ the Gumbel-Softmax function on volume density as a continuous and differentiable approximation. First, we calculate a soft probability $p_i$ for the $i$-th object branch as 
\begin{equation}
\setlength\abovedisplayskip{5pt}
\begin{aligned}
p_i = \frac{\exp^{((\sigma_{\text{object}}^{i}+g_i)/tau)}} {\sum_j \exp^{((\sigma_{\text{object}}^{j}+g_j)/tau)}}
\end{aligned} \label{soft_p}
\setlength\belowdisplayskip{5pt}
\end{equation}
The above $g_1,g_2,..,g_j$ are the independent and identically distributed samples drawn from $\operatorname{Gumbel}(0,1)^1$~\cite{DBLP:conf/iclr/JangGP17}. The symbol $tau$ is the temperature ratio set to 0.1 in our experiments. 
Then, we can obtain one-hot composition weights from $p_i$ for combining the predictions of all the object/background branches, which ensures one-hot activation for all the object/background radiance fields, and enables effective backpropagation for learning object-specific radiance fields and their composition.
Specifically, we calculate the one-hot weights $\mathbf{M}_{{obj}}$ as Eq.~\ref{gumbel_mask}. 
Let us denote $\mathbf{P}$ as the soft probability vector, containing $K+1$ predictions calculated by Eq.~\ref{soft_p}. 
$\mathbf{M}_{{obj}}$ is a one-hot vector of the $K+1$ elements. 
The one-hot vector is calculated with a stop gradient operation $f_{sp}$ on $\mathbf{P}$, which not only reserves the one-hot activation but also allows the gradient backpropagated through $\mathbf{P}$. Then, the computation of $\mathbf{M}_{obj}$ is written as:

\begin{equation}
\setlength\abovedisplayskip{8pt}
\mathbf{M}_{{obj}}= f_{\textrm{h}}\big(\underset{i}{\arg \max }[g_i+\log \sigma_{\text{object}}^{i}]\big) 
 - f_{sp}(\mathbf{P}) + \mathbf{P},
\label{gumbel_mask}
\setlength\belowdisplayskip{4pt}
\end{equation}
where $f_{\textrm{h}}(\cdot)$ denotes a one-hot function; Then, we obtain a composed scene radiance field $(\mathbf{c}_\text{comp}, \mathbf{\sigma}_\text{comp})$ of all decomposed branches $\{\mathbf{c}_\text{object}^k, \mathbf{\sigma}_\text{object}^k\}_{k=1}^{K+1}$, by multiplying the above one-hot weights as follows:

\begin{equation}
\setlength\abovedisplayskip{0pt}
\mathbf{c}_{\text{comp}} = \sum_{K+1} \mathbf{M}_{{obj}}^k \mathbf{c}_{\text{object}}^k, \ \ 
\sigma_{\text{comp}} = \sum_{K+1} \mathbf{M}_{{obj}}^k \sigma_{\text{object}}^k,
\setlength\belowdisplayskip{6pt}
\end{equation}
Our novel one-hot object radiance activation can not only ensure a reasonable composition of all the object radiance fields but also force that only one object volume density prediction can be activated in the gradient backpropagation, which can also effectively help the object/background decomposition. 

\begin{figure*}[!tp]
    \centering    \includegraphics[width=0.90\linewidth]{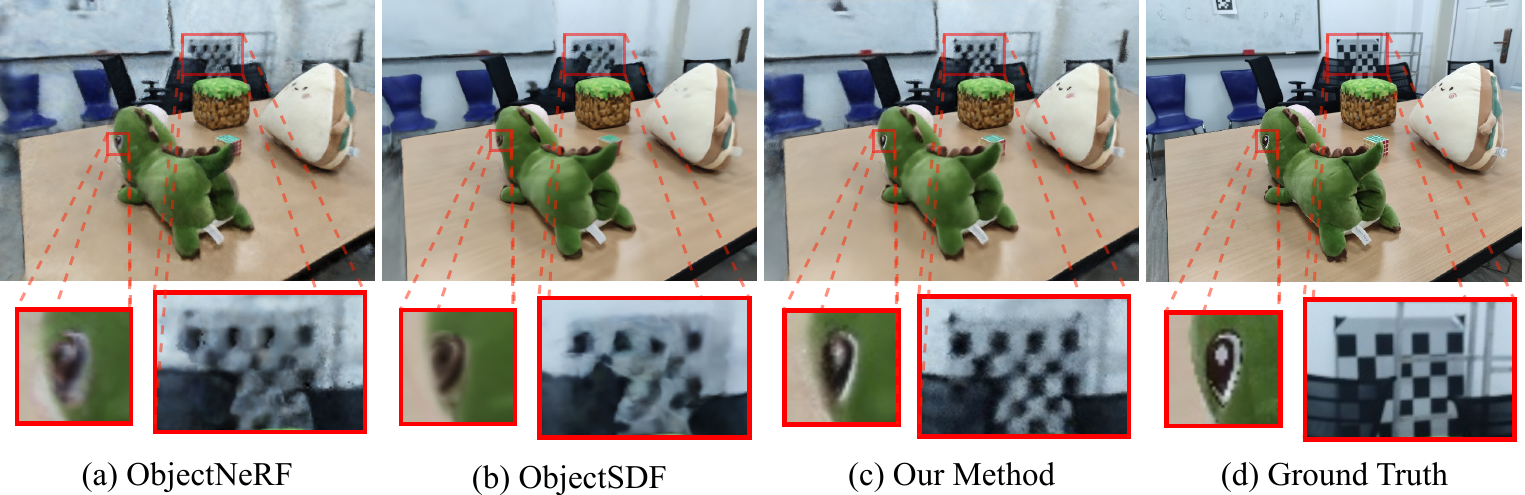}
    \vspace{-2pt}
    \caption{Qualitative comparison of novel view synthesis on the ToyDesk dataset. We compared with ObjectNeRF~\cite{yang2021objectnerf} and ObjectSDF~\cite{wu2022objectsdf}. The results show that the rendered images of our framework achieve better generation quality for both the foreground objects (\textit{e.g.}, the green dinosaur) and the background (\textit{e.g.}, the chessboard). }
    \vspace{-15pt}
    \label{fig:nvs1}
\end{figure*}

\par\noindent\textbf{Residual radiance field composition.}
After the one-hot object radiance activation, we further combine the composited radiance fields from objects/background branches with the predicted residual radiance field, to produce a final composited scene radiance field, using point-wise average pooling operation applied on the color and density. 
Then we apply a color loss $\mathcal{L}_{\text{comp}}$ for the pixels rendered from the composited radiance field. 

\subsection{Optimization and Inference}

\noindent\textbf{Overall framework optimization.} The overall optimization loss for training the joint framework consists of the following four different loss functions: 
\begin{equation}
\begin{aligned}
\mathcal{L}_{\text{overall}} =  \mathcal{L}_{\text{comp}} +   \lambda_1 \mathcal{L}_{\text{object}}
 + \lambda_2 \mathcal{L}_{\text{color\_pseudo}} + \lambda_3 \mathcal{L}_{\text{one-hot}}, 
\end{aligned}
\end{equation}
where, $\lambda_1$, $\lambda_2$, and $\lambda_3$ are the weights for balancing the learning with the different loss functions.

\par\noindent\textbf{Novel View Rendering and Scene Editing}
After training the joint framework, we can render images of arbitrary perspectives based on the composited radiance field. By re-organizing the object radiance fields, we can generate images with object manipulations, including removal, addition, duplication, and changing position (e.g., rotation).

%% file: 04_experiments.tex
\section{Experiments}
\label{sec:experients}

We aim at performing joint scene novel view synthesis and editing based on our proposed unified decompositional and compositional NeRF framework. We thus evaluate our method on both tasks using challenging public benchmarks.

\begin{table}[!t]
    \centering
    \resizebox{\columnwidth}{!}{
    \begin{tabular}{lccccccc}
    \toprule \multirow{2}{*}{ Methods } & \multicolumn{3}{c}{ ToyDesk2 } \\
    \cline { 2 - 4 } & PSNR $\uparrow$ & SSIM $\uparrow$ & LPIPS $\downarrow$ \\
    \toprule Our Full-Model & \pmb{25.7560} & \pmb{0.8126} & \pmb{0.4488} \\
    \midrule 
    w/o 3D One-hot Density Regularization & 25.7362 & 0.8116 & 0.4524 \\
    w/o Gumbel-Softmax Activation & 22.9972 & 0.7217 & 0.5452 \\ 
    w/o 2D In-painting pseudo Supervision & 25.6881 & 0.8127 & 0.4516 \\
    w/o Residual Radiance Composition & 25.2288 & 0.8003 & 0.4589 \\
    \bottomrule 
    \end{tabular}
    }
    \vspace{-2pt}
    \caption{
    Quantitative comparison of different variants of the proposed unified decompositional and compositional NeRF framework on the ToyDesk2.}
    \vspace{-18pt}
    \label{tab:abl}
\end{table}

\subsection{Experimental Setup}

\begin{figure}[!tp]
    \centering
    \includegraphics[width=\linewidth]{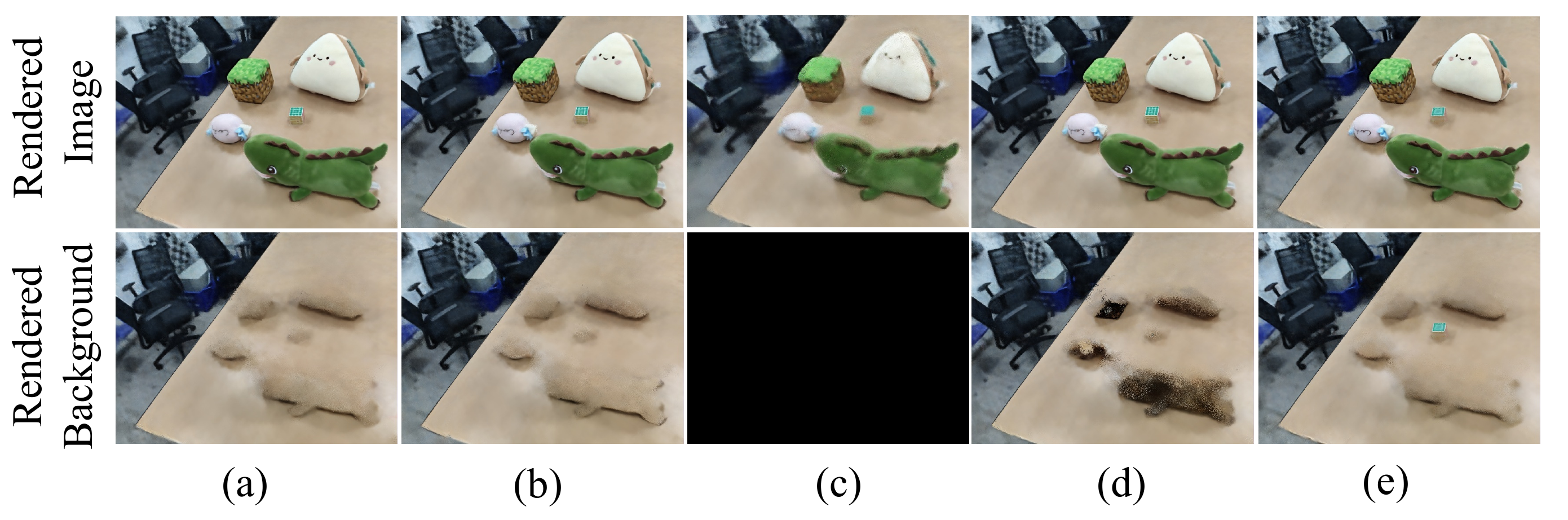}
    \vspace{-15pt}
    \caption{Qualitative results of the ablation study: (a) Our full model; {(b) w/o 3D One-hot Density Regularization; (c) w/o Gumbel-Softmax Activation;} (d) w/o 2D In-painting pseudo Supervision. (e) w/o Residual Radiance Composition. } 
    \vspace{-18pt}
    \label{fig:abl}
\end{figure} 

\par\noindent\textbf{Datasets.} We use ToyDesk ~\cite{yang2021objectnerf} and ScanNet ~\cite{dai2017scannet} datasets for the evaluation. \textbf{ToyDesk} consists of two sets of indoor multi-view images. In each scene, several toys are put on a desk in the center of the room. The multi-view images are captured by looking at the desk center with 360 degrees around the desk, and the camera poses and meshes are provided. The instance semantic labels are annotated on object meshes and then projected onto each perspective to obtain their corresponding 2D semantic label maps. \textbf{ScanNet} is an RGB-D video dataset that contains more than 1500 scans of indoor scenes for multiple 3D understanding tasks. Handheld RGB-D scanners are used for scanning various indoor scenes. Rich annotations such as camera pose, semantic segmentation, and surface reconstruction are also provided. Following ObjectNeRF~\cite{yang2021objectnerf}, we choose four scenes, \textit{i.e.}, 0024\_00, 0038\_00, 0113\_00, and 0192\_00, from the whole dataset for the experiment. 

\par\noindent\textbf{Implementation details.} Our joint framework consists of a shared backbone, background head, object head, and guidance head. The share backbone is a 4-layer Multi-Layer Perceptron (MLP), and the output generic implicit scene feature has a dimension of 256. The follow-up heads are all three-layer MLPs with a color and density prediction layer to generate corresponding radiance fields. Note that our one-hot regularization and activation module has no parameters to be optimized. 
We also utilize positional encoding to increase the representation ability for high-frequency details. Specifically, we adopt ten levels for coordinates and four levels for ray directions. 
The framework is trained end-to-end from scratch, without voxel feature initialization as in~\cite{yang2021objectnerf} or geometric initialization as in~\cite{wu2022objectsdf}.

\begin{figure*}[!tp]
    \centering
    \includegraphics[width=0.90\linewidth]{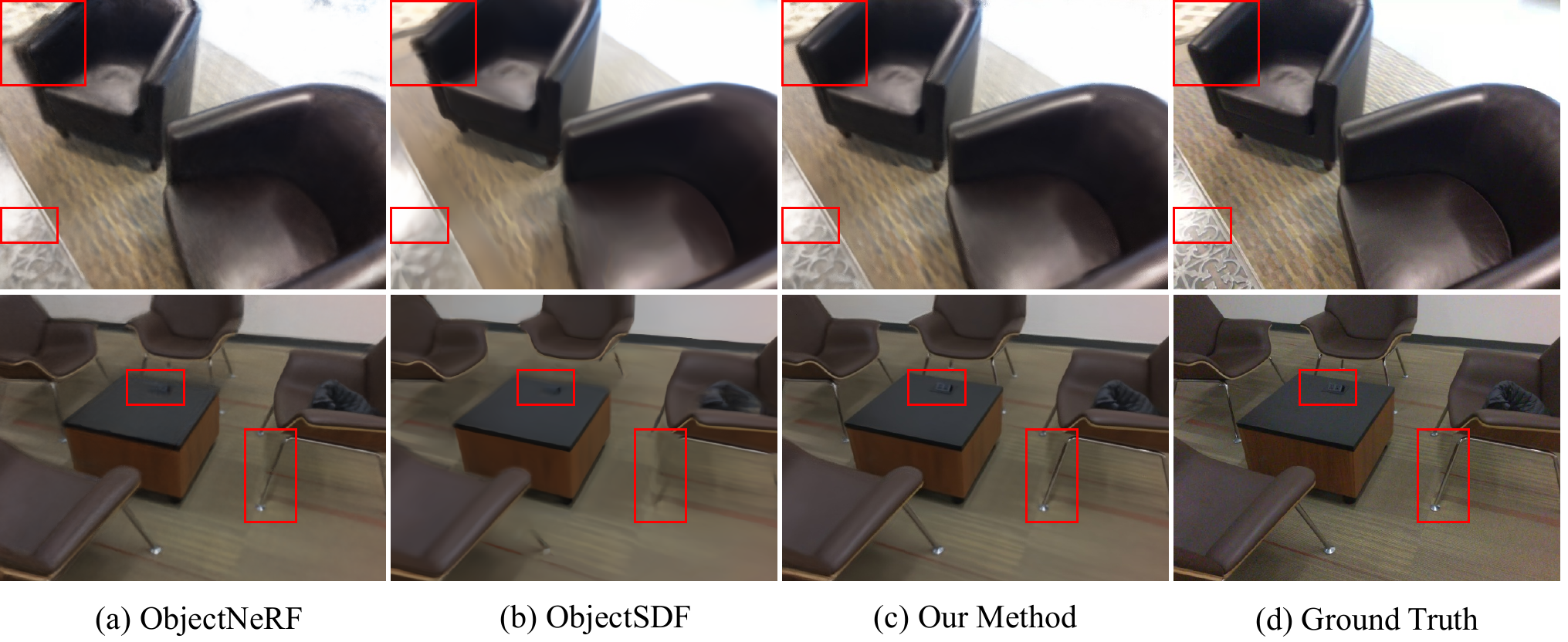}
    \vspace{-3pt}
    \caption{Qualitative comparison of novel view synthesis on the ScanNet dataset. Compared with ObjectNeRF~\cite{yang2021objectnerf} and ObjectSDF~\cite{wu2022objectsdf}, our results also achieve better fidelity. We highlight the details in the red boxes. } 
    \vspace{-15pt}
    \label{fig:nvs2}
\end{figure*}

\begin{figure}[!tp]
    \centering
    \includegraphics[width=0.85\linewidth]{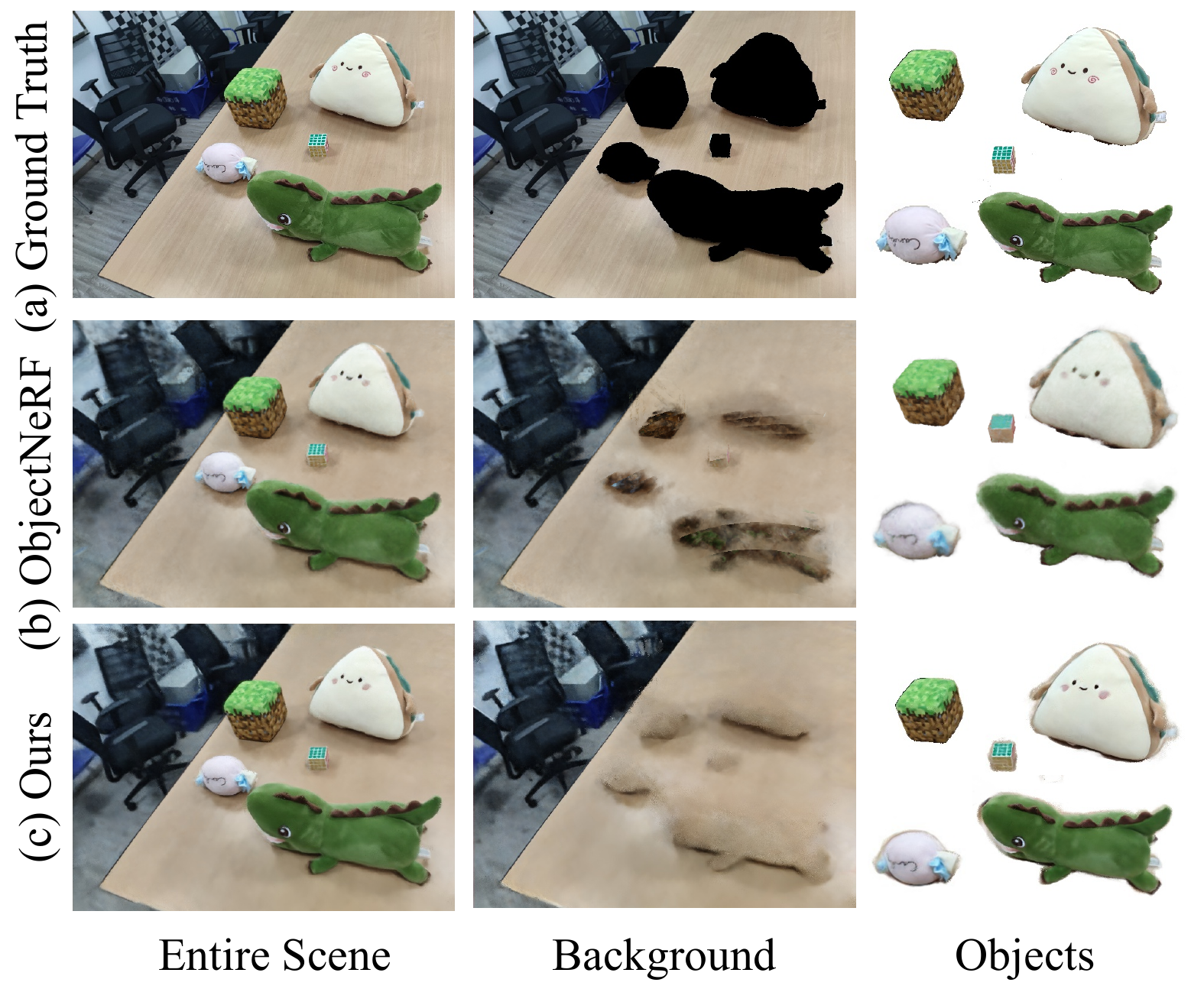}
    \vspace{-2pt}
    \caption{The Decomposition results on the ToyDesk dataset compared with ObjectNeRF~\cite{yang2021objectnerf}. We show the rendered results of our composited radiance field (the first column) and each background/object radiance field (the second and third columns). } 
    \label{fig:decomp1}
    \vspace{-18pt}
\end{figure} 

\subsection{Model Analysis}
\par\noindent\textbf{Baseline Models.} 
To verify the effectiveness of our designs for scene decomposition and composition, We conduct several ablation studies considering different variants: 
(i) 'w/o 3D One-hot Density Regularization' means that we remove the one-hot regularization term for the uncertain regions in the background radiance field. 
(ii) 'w/o Gumbel-Softmax Activation' indicates that we do not use the Gumbel-Softmax one-hot weight for compositing the object radiance fields. 
(iii) 'w/o In-painting Supervision' means that we do not apply the in-painting pseudo supervision for the background radiance field. 
(iv) 'w/o Residual Radiance Composition' means that we do not use the guidance radiance field for fine sampling and remove the residual radiance composition in the fine stage.

\par\noindent\textbf{Effect of 3D one-hot density regularization.} 
The quantitative results based on different evaluation metrics, \textit{i.e.}, PSNR, SSIM, and LPIPS, are shown in Table~\ref{tab:abl}. Several examples of rendered images from the learned background radiance field are presented in Fig.~\ref{fig:abl}. We evaluate the effectiveness of 3D one-hot density regularization by disabling the loss term. As can be observed from Table~\ref{tab:abl}, our joint full mode with 3D one-hot density regularization achieves better results than the model without it.

\begin{table}[!tp]
\begin{center}
\resizebox{0.48\textwidth}{!}{
\begin{tabular}{c||l||ccc}
\toprule[1.0pt]
Dataset & \multicolumn{1}{c||}{Methods} & PSNR $\uparrow$ & SSIM $\uparrow$ & LPIPS$\downarrow$ \\ 
\midrule[1.0pt]
\multirow{3}{*}{\begin{tabular}{l} ToyDesk \end{tabular}}
~ & Object-NeRF~\cite{yang2021objectnerf} & 22.1736 & 0.7723 & 0.4711 \\
~ & Object-SDF~\cite{wu2022objectsdf}  & 22.0298 & -      & -      \\
~ & Our Method  & \pmb{23.1435} & \pmb{0.7899} & \pmb{0.4691} \\
\midrule
\multirow{3}{*}{\begin{tabular}{l} ScanNet \end{tabular}}
~ & Object-NeRF~\cite{yang2021objectnerf} & 25.2640 & 0.8047 & 0.4094 \\
~ & Object-SDF~\cite{wu2022objectsdf}  & 25.2325 & -      & -      \\
~ & Our Method  & \pmb{26.1349} & \pmb{0.8249} & \pmb{0.3951} \\
\bottomrule[1.0pt]
\end{tabular}
}
\end{center}
\vspace{-6pt}
\caption{Quantitative comparison on novel view synthesis. We compared with ObjectNeRF~\cite{yang2021objectnerf} and ObjectSDF~\cite{wu2022objectsdf}. ‘-’ indicates the metrics are not reported by the authors in the paper. The evaluation results show that our synthesis results quantitatively outperform these two SOTA methods on all the shown metrics. }
\label{tab:nvs}
\vspace{-16pt}
\end{table}

\par\noindent\textbf{Effect of Gumbel-Softmax object radiance activation.} We also investigate the effect of using Gumbel-Softmax one-hot activation for the composition of background/object radiance fields. We use an alternative composition method, in which all the local object/background radiance fields that contribute to the same 3D points in the composited field, are directly added instead of using the Gumbel-Softmax one-hot activation weights, as introduced in the method part.
{As shown in Fig.~\ref{fig:abl}c, without the Gumbel-Softmax one-hot activation, the model cannot successfully learn a decomposed background radiance field (all black), confirming our motivation of using the Gumbel-Softmax activation for a more effective scene representation in a unified framework. }

\par\noindent\textbf{Effect of 2D in-painting pseudo supervision and residual radiance composition.} We also show the contribution of the proposed 2D in-painting pseudo-supervision. If we directly disable the pseudo supervision in the framework, the qualitative results are slightly lower, as shown in the next to last row in Table.~\ref{tab:abl}. At the same time, as can be seen in Fig.~\ref{fig:abl}, lacking pseudo supervision suffers from significantly downgraded extrapolation in the region occupied by the foreground objects in the 3D space, which indicates that the proposed scheme can effectively help the local object/background disentanglement, especially for learning the radiance field representation of the background on uncertain regions. As shown in Table.~\ref{tab:abl} and Fig.~\ref{fig:abl}, if we remove the residual radiance composition,  the overall qualitative metrics and background decomposition both decline, which demonstrates the importance of using a global radiation field to provide coarse sampling guidance and also further merged as residual information in the fine stage.

\begin{figure}[!tp]
    \centering
    \includegraphics[width=0.80\linewidth]{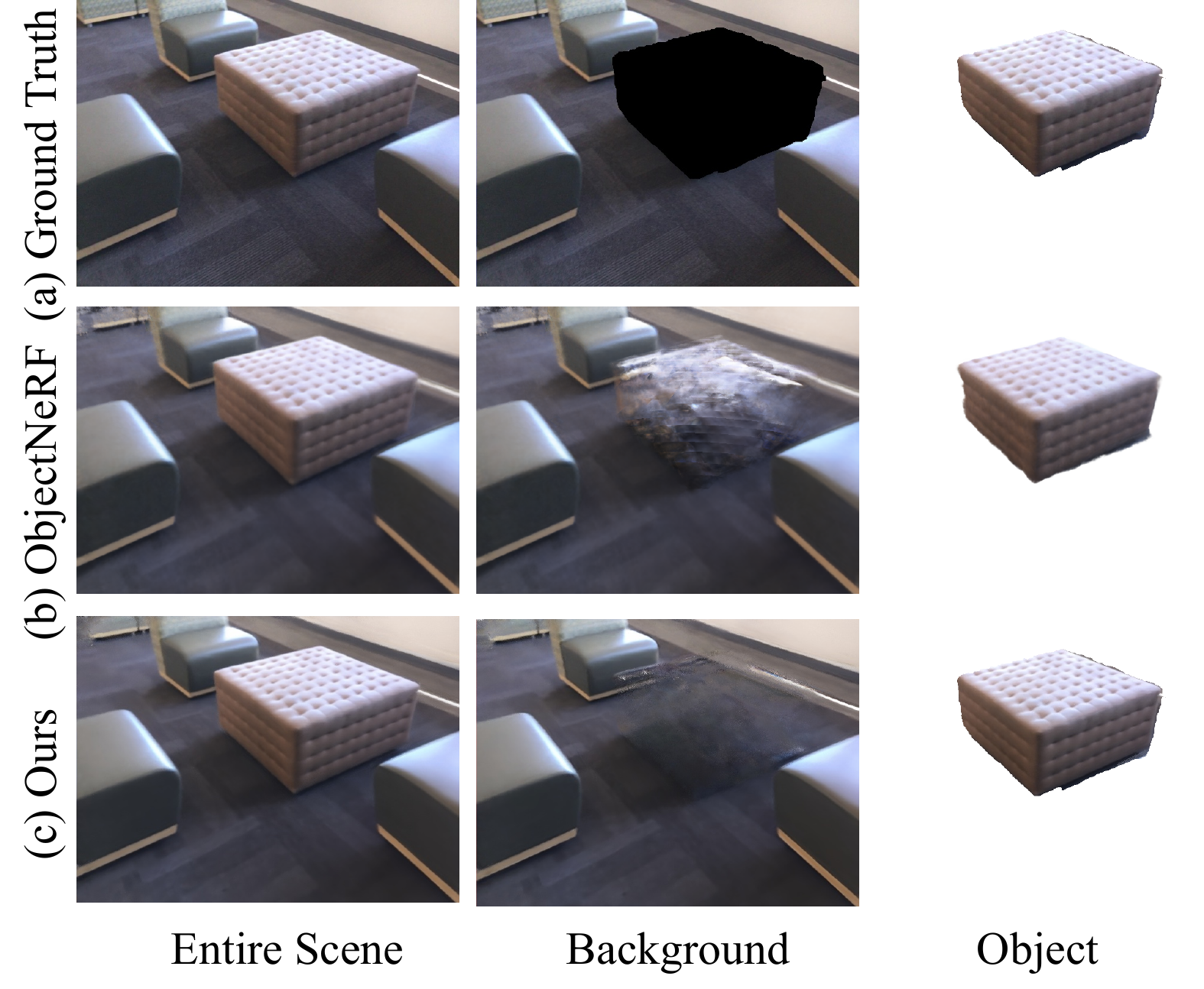}
    \vspace{-2pt}
    \caption{The Decomposition results on the ScanNet dataset compared with ObjectNeRF~\cite{yang2021objectnerf}. Our background has fewer artifacts when removing the foreground object.} 
    \label{fig:decomp2}
    \vspace{-15pt}
\end{figure}

\subsection{Results on Novel View Synthesis}
We compare our methods on the task of novel view synthesis with two state-of-the-art methods, \textit{i.e.}, ObjectNeRF~\cite{yang2021objectnerf} and ObjectSDF~\cite{wu2022objectsdf}. Our train-test split follows ObjectNeRF~\cite{yang2021objectnerf}. 
We measure the quality of synthesized images on the evaluation metrics of PSNR, SSIM, and LPIPS. In detail, for our joint framework, we use the finally composited entire scene radiance field for the image rendering. For ObjectNeRF and ObjectSDF on the Toydesk dataset, we use the rendered images provided by the authors for comparison. ObjectSDF adopts a different train-test split and label mapping strategy on ScanNet, so we re-train the ObjectSDF on ScanNet following our setting. 

In Table~\ref{tab:nvs}, our synthesis results quantitatively outperform the comparison methods on all the shown metrics. We also present the image examples rendered from novel views in Fig.~\ref{fig:nvs1} and Fig.~\ref{fig:nvs2}. 
As can be observed in Fig.~\ref{fig:nvs1}, the rendered images of our framework achieve better generation quality for both the foreground objects (\textit{e.g.}, the green dinosaur) and the background (\textit{e.g.}, the chessboard) than both ObjectNeRF~\cite{yang2021objectnerf} and ObjectSDF~\cite{wu2022objectsdf}, proving the capability of our method in modeling both global scene consistency and local scene details. Fig.~\ref{fig:nvs2} shows the synthesized images on ScanNet. We can also observe that our results outperform others in the overall fidelity and local details marked in the red boxes.

\subsection{Results of Object Decomposition and Editing}
We also conduct experiments to show the decomposition effectiveness of our method, especially for the background radiance field. As shown in Fig.~\ref{fig:decomp1} and Fig.~\ref{fig:decomp2}, we rendered the composited radiance field and each local object/background radiance field to show the decomposition results. 
The background images in Fig.~\ref{fig:decomp1} clearly show that our method can generate a clearer table without unsatisfactory artifacts such as strange colors and black shadow effects when removing all the foreground objects on the table. Although with some remained black contours, the rendered pixel colors of the regions below the objects are significantly more reasonable compared to ObjectNeRF~\cite{yang2021objectnerf}. Note that the region under the toys is unseen in any view. 
Our decomposition module enables clear background and object-specific details, while the composition design helps the rendered entire scene achieve higher quality than the baseline.
In Fig.~\ref{fig:decomp2}, we further show that our pipeline can also provide better background rendering when removing foreground objects on the images from the ScanNet dataset. 
The better decomposition naturally benefits downstream object editing tasks. 
Fig.~\ref{fig:edit} shows that, after learning composited radiance fields, our method can perform rotation and duplication for the white triangle and candy by re-compositing the object and background radiance field representations. We add more editing results in the supplementary materials.

%% file: 10_conclusion.tex
\begin{figure}[!tp]
    \centering
    \includegraphics[width=0.90\linewidth]{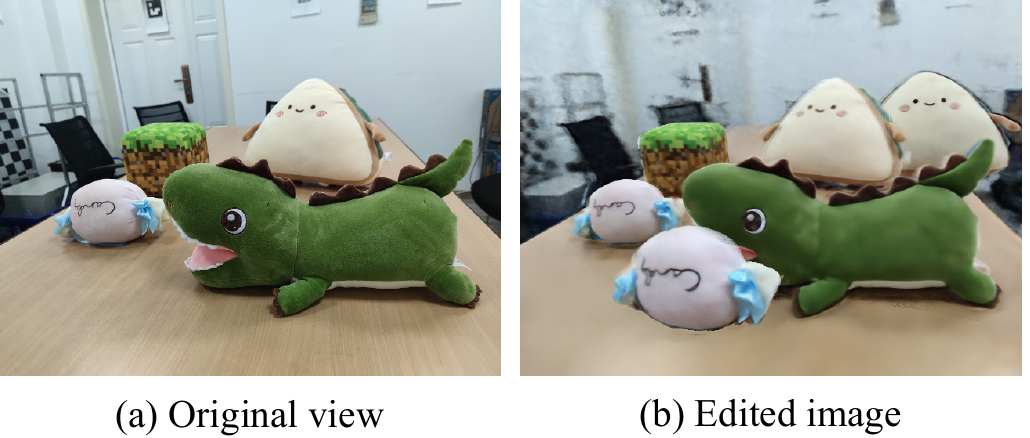}
    \vspace{-2pt}
    \caption{Illustration of the editing results. (a) shows the image in the original view, and (b) shows the editing result from the same perspective. We do the rotation and duplication for the white triangle and candy by re-composting the learned radiance fields. } 
    \vspace{-16pt}
    \label{fig:edit}
\end{figure}

\section{Conclusion}
\vspace{-5pt}
\label{sec:conclusion}
We propose a unified NeRF framework with a novel decomposition and composition design to jointly model real-world scenes with both global and local implicit representations. To solve the ambiguous extrapolation problem for unseen areas, we also propose a 3D one-hot density regularization strategy and a 2D in-painting pseudo supervision. 
Our experimental evaluation demonstrates that our method outperforms the state-of-the-art object-compositional methods for both novel-view synthesis and scene editing tasks.

%% file: 12_appendix.tex
\appendix
\label{sec:appendix}

In this supplementary material, we introduce the model architecture in Sec.~\ref{sec:arch}, the in-painting details in Sec.~\ref{sec:lama}, more experimental results in Sec.~\ref{sec:exp}, and further elaboration on joint scene decomposition and composition in Sec.~\ref{sec:ela}

\section{Model Architecture}
\label{sec:arch}
The model architecture is shown in Fig.~\ref{fig:arch}. Following NeRF~\cite{mildenhall2020nerf} and ObjectNeRF~\cite{yang2021objectnerf}, we utilize a 4-layer Multi-Layer Perceptron (MLP) as the shared backbone for learning generic implicit representations of scene points. The implicit point representations are then passed to specific heads for predicting object/background radiance fields. We use three MLPs for the density prediction and five MLPs for the color prediction. In the fine stage, we use three separate heads with the same structure for the background, object, and guidance prediction head. In the coarse stage, only the guidance prediction head is used.  
We set the positional embedding frequency to 10 for the input coordinates and 4 for the input ray direction. 

\section{Implementation of In-painting Process}
\label{sec:lama}
To learn a better background-specific radiance field, especially the extrapolation on the occluded region, we involve a pre-trained in-painting model lama~\cite{suvorov2021resolution} to fill the uncertain occluded regions on the 2D images. 
Specifically, as shown in Fig.~\ref{fig:lama}, we pass the original images with their corresponding foreground object masks to the pre-trained lama model. Then we use the in-painted pixel colors of the background region as pseudo color supervision. To mitigate imprecise mask contours, we apply an appropriate erosion for the foreground masks before passing the images to the lama model.

\section{More Experimental Results}
\label{sec:exp}

\subsection{More Results on Scene Rendering}
We show more qualitative scene rendering results on the ScanNet dataset in Fig.~\ref{fig:nvs}. Compared with other state-of-the-art methods ObjectNeRF~\cite{yang2021objectnerf} and ObjectSDF~\cite{wu2022objectsdf}, our rendered images produce more fine-grained details on the overall-view quality and object details. 
Note that for the scene rendering comparison, the test-set images are directly provided by the authors of ObjectNeRF~\cite{yang2021objectnerf}, which is the same as their paper, while different from the train-test split of their open-sourced code.

\subsection{More Results on Scene Editing}
We show more qualitative scene editing results on both ToyDesk and ScanNet datasets in Fig.~\ref{fig:edit_supp}. As we can see, by re-organizing the object radiance fields, we can generate images with object manipulations, including removal, duplication, and changing object position (\textit{e.g.}, the movement and rotation). 

\begin{figure}[!bp]
    \vspace{-50pt}
    \centering    
    \includegraphics[width=0.9\linewidth]
    {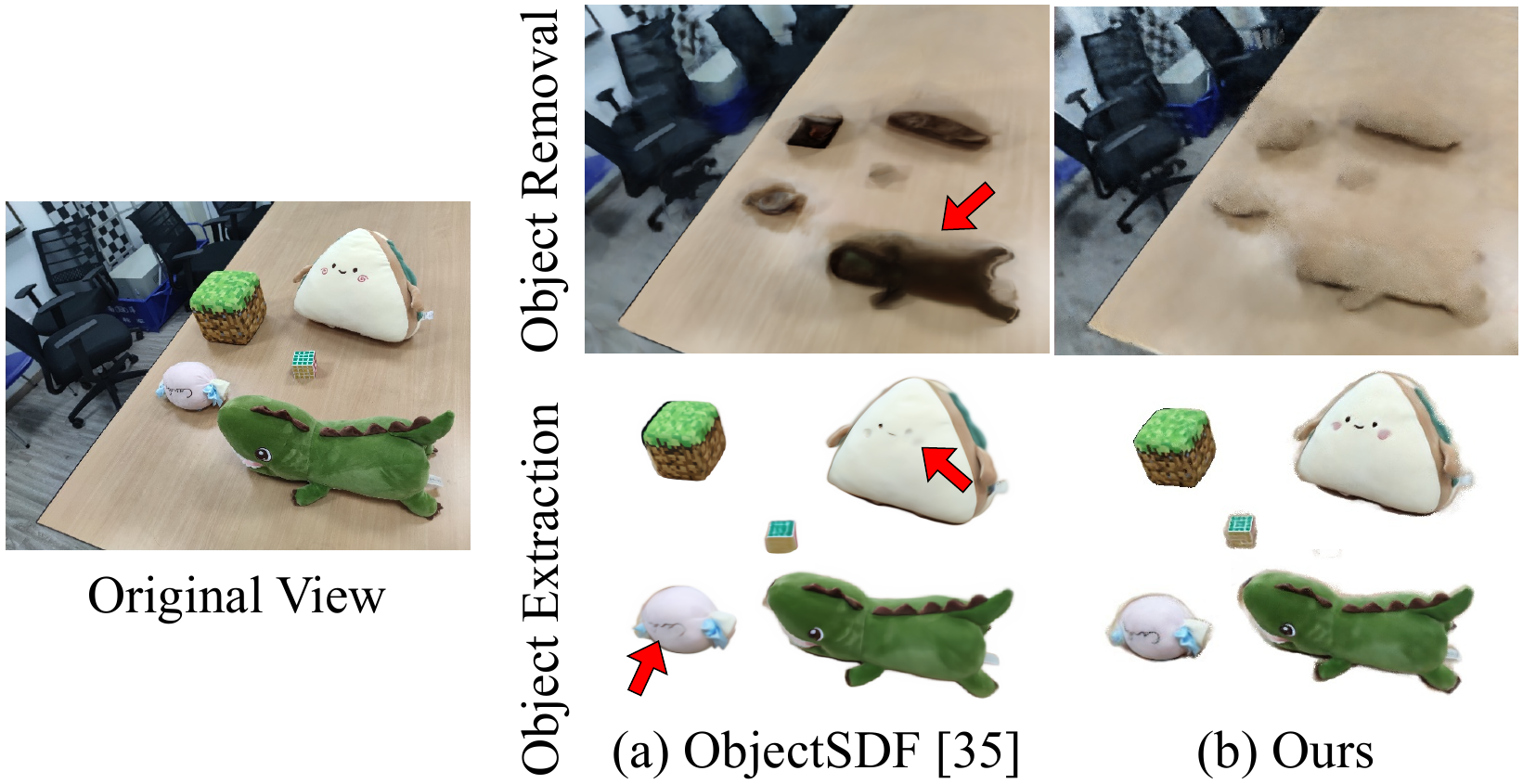}
    \caption{{The scene editing comparison with ObjectSDF~\cite{wu2022objectsdf}, using editing operations of both object removal and object extraction for the scene decomposition on ToyDesk2.}}
    \label{fig:rebuttal_decomp}
\end{figure}

\begin{figure*}[tbp]
    \centering
    \includegraphics[width=\linewidth]{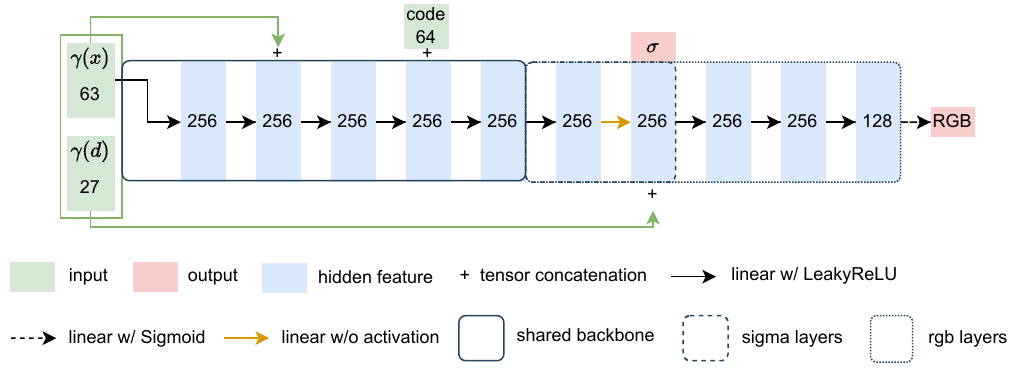}
    \caption{Illustration of the detailed model architecture. We set the positional embedding frequency to 10 for the input coordinates $x$ and 4 for the input ray direction $d$, so the input embedding dimensions of these two are $3+3\times2\times10=63$ and $3+3\times2\times4=27$, respectively. The volume density $\sigma$ and RGB color are predicted by specific layers on the top of the shared backbone. We use three separate heads with same structure for the background, object and guidance prediction. } 
    \label{fig:arch}
\end{figure*}

\begin{figure*}[tbp]
    \centering
    \includegraphics[width=\linewidth]{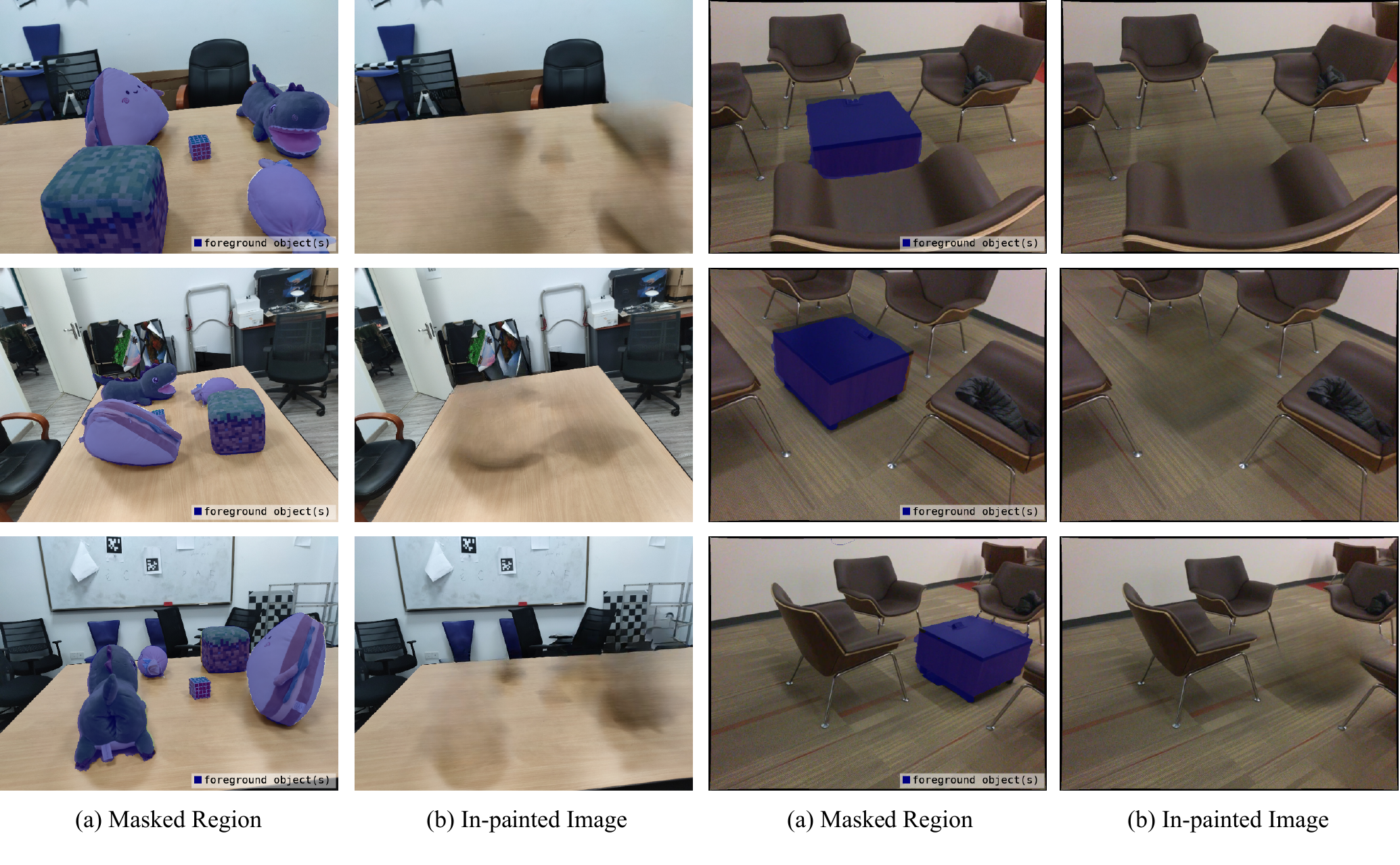}
    \caption{Illustration of the In-painting process. (a) shows the original image and foreground object masks, while (b) shows the in-painted images from the pre-trained lama model. The uncertain occluded regions are filled with pseudo colors. Note that the 2D in-painting may bring new ambiguity for the regions seen in other views. For example, the lama model does not correctly in-paint the armrest (in row 1, column 2) and the chair leg (in row 3, column 4). However, supervision from other views can mitigate most of the ambiguities. }
    \label{fig:lama}
\end{figure*}

\begin{figure*}[tbp]
    \centering
    \includegraphics[width=\linewidth]{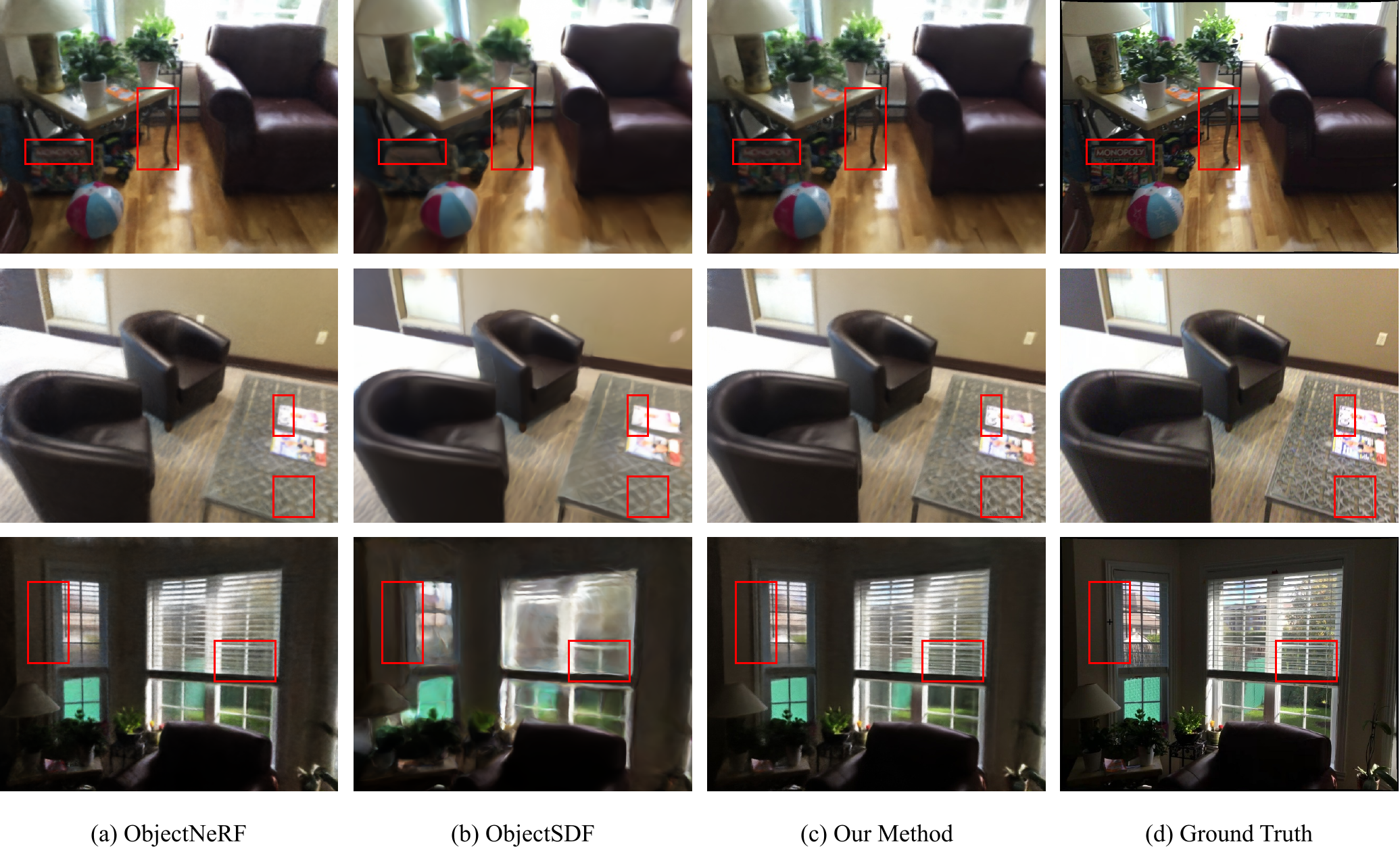}
    \caption{More qualitative scene rendering results on the ScanNet dataset. We highlight the details in the red boxes. Our rendering results outperform other SoTA methods ObjectNeRF~\cite{yang2021objectnerf} and ObjectSDF~\cite{wu2022objectsdf} with higher overall-view quality and more object details.}
    \label{fig:nvs}
\end{figure*}

\begin{figure*}[tbp]
    \centering
    \includegraphics[width=\linewidth]{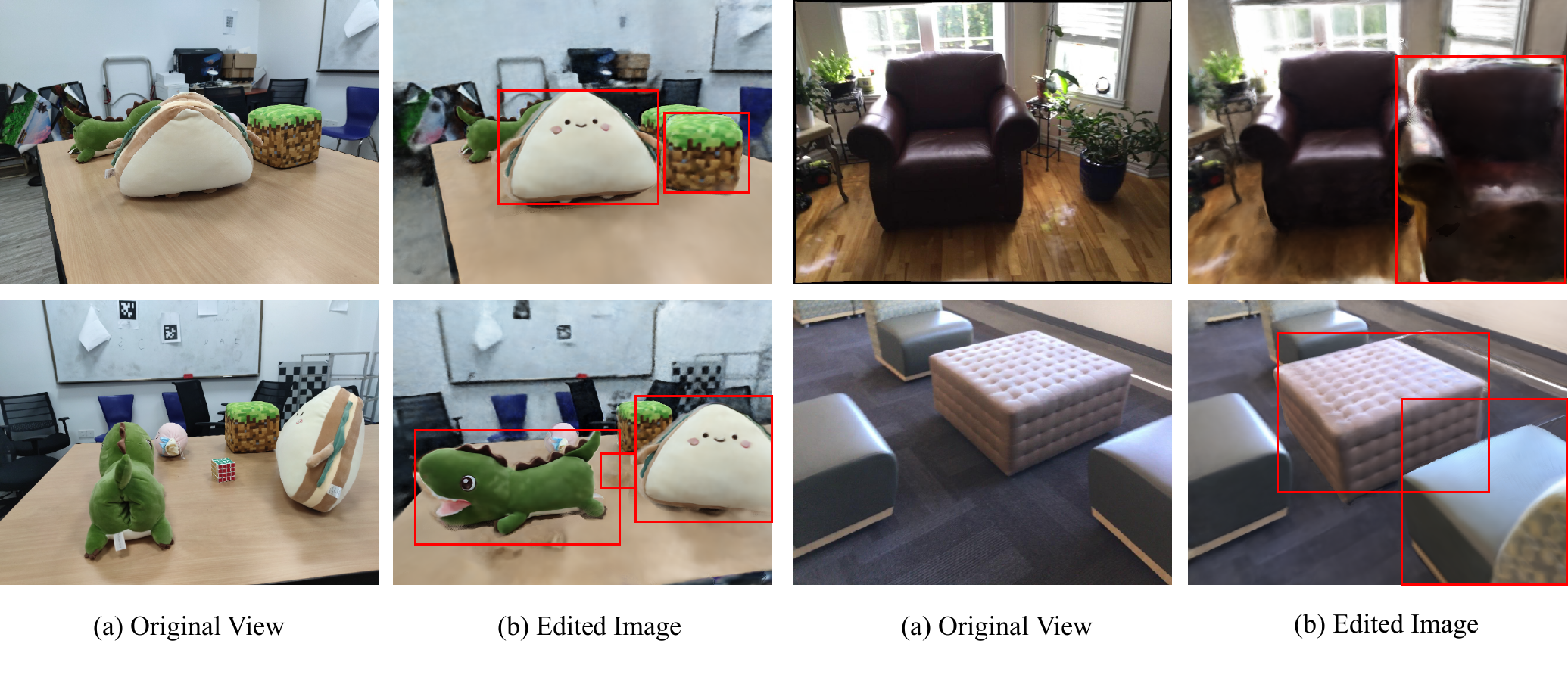}
    \caption{More qualitative scene editing results on both ToyDesk and ScanNet datasets. We highlight the editings in the red boxes, including duplication (the green cube, triangle toy, and the sofa), removal (the Rubik's cube), rotation (the dinosaur and the triangle toy), and movement (the sofa and the mattress). }
    \label{fig:edit_supp}
\end{figure*}

{
ObjectSDF~\cite{wu2022objectsdf} focuses on implicit object geometry modeling.  
Although ObjectSDF mentioned that scene editing can be a potential application, it did not address the object editing problem in their paper or provide any open-sourced code for editing. 
In contrast to them, we focus on editable novel view synthesis, where novel view scene rendering and editing are both our important objectives. 
In Fig.~\ref{fig:rebuttal_decomp}, we conducted a decomposition comparison with ObjectSDF~\cite{wu2022objectsdf}, using two editing operations, \emph{i.e.}, object removal and object extraction. As highlighted by the red arrows, our model can achieve finer-grained object-level rendering results, 
with a clearer background decomposition (see black shadows on the table).  
}

\subsection{Video Demos on Scene Editing}
We also upload one editing video demo, which can provide more qualitative results to directly compare with ObjectNeRF~\cite{yang2021objectnerf}. The video demos can clearly show that our model generates superior editing results compared to ObjectNeRF~\cite{yang2021objectnerf}, with better video quality and background decomposition. 

{\section{Further elaboration on joint scene decomposition and composition}
~\label{sec:ela}
Our framework targets joint scene editing and synthesis. The scene decomposition can provide the capability of learning disentangled representations of different background/objects, allowing for scene editing, while scene composition learns an entire scene representation for novel view synthesis. The decomposition and composition are jointly modeled and are technically correlated in our unified framework. As shown in Fig.~1 (paper), the forward pass of the network composes the background/objects into the whole scene, while the gradient backward pass can assist in the decomposition process. 
Thus, these two can be united to facilitate the consistency constraint in the unified optimization framework. We will add the discussion on this point in the revision. }

